\documentclass[letterpaper, 10pt, conference]{IEEEtran}
\IEEEoverridecommandlockouts

\usepackage[T1]{fontenc}
\usepackage[utf8]{inputenc}
\usepackage{amsmath,amssymb,amsfonts}
\usepackage{graphicx}
\usepackage{booktabs}
\usepackage{enumitem}
\usepackage{cite}
\usepackage{hyperref}
\usepackage{xcolor}
\usepackage{tikz}
\usetikzlibrary{shapes,arrows,positioning,backgrounds,fit,calc,shadows.blur}
\usepackage{multirow}
\usepackage{tabularx}
\usepackage{balance}

\newcommand{\vishnu}[1]{}
\newcommand{\hari}[1]{}
\newcommand{\miheer}[1]{}

\title{\LARGE \bf Singularity Avoidance in Inverse Kinematics: \\
A Unified Treatment of Classical and Learning-based Methods}

\author{%
  \IEEEauthorblockN{Vishnu Rudrasamudram\textsuperscript{*} and Hariharasudan Malaichamee\textsuperscript{*}}
  \thanks{\textsuperscript{*}Vishnu Rudrasamudram and Hariharasudan Malaichamee contributed equally to this work.}
  \thanks{Emails: vishnurs@ieee.org, mhariharasudan@gmail.com}
}

\begin{document}
\maketitle

% ============================================================================
\begin{abstract}
Singular configurations cause loss of task-space mobility, unbounded joint velocities, and solver divergence in inverse kinematics (IK) for serial manipulators. No existing survey bridges classical singularity-robust IK with rapidly growing learning-based approaches. We provide a unified treatment spanning Jacobian regularization, Riemannian manipulability tracking, constrained optimization, and modern data-driven paradigms. A systematic taxonomy classifies methods by retained geometric structure and robustness guarantees (formal vs.\ empirical). We address a critical evaluation gap by proposing a benchmarking protocol and presenting experimental results: 12~IK solvers are evaluated on the Franka Panda under position-only IK across four complementary panels measuring error degradation by condition number, velocity amplification, out-of-distribution robustness, and computational cost. Results show that pure learning methods fail even on well-conditioned targets (MLP: 0\% success, ${\sim}$10\,mm mean error), while hybrid warm-start architectures - IKFlow (59\%$\to$100\%), CycleIK (0\%$\to$98.6\%), GGIK (0\%$\to$100\%) - rescue learned solvers via classical refinement, with DLS converging from initial errors up to 207\,mm. Deeper singularity-regime evaluation is identified as immediate future work.
\end{abstract}

% ============================================================================
\section{Introduction}
\label{sec:intro}

Inverse kinematics (IK) is a core problem in robot motion and control: given a desired end-effector pose, determine the joint configuration or joint motion required to realize it. For most serial manipulators, the forward kinematic mapping is nonlinear and does not admit a globally convenient closed-form inverse, especially when additional objectives such as joint-limit avoidance, obstacle avoidance, or task hierarchies are imposed~\cite{nakamura1987_task_priority, siciliano1991_redundancy, sentis2005_wholebodycontrol}. IK is therefore most often solved numerically through inverse differential kinematics, where singular configurations become a dominant failure mode: they cause loss of instantaneous mobility in task-space directions while amplifying joint velocities, leading to oscillatory behavior, degraded tracking, and solver divergence~\cite{wampler1986_dls, duleba2023_singularity_robust, chiaverini1994_dls_review, nielsen1991_singularity_throughput, chiaverini1997_singularity_robust}. These effects are especially severe in safety-critical applications---industrial manipulators, surgical platforms, collaborative robots---where unbounded or poorly conditioned joint commands are unacceptable.

Over the last several decades, singularity handling in IK has evolved from classical Jacobian regularization methods into a broader design space that includes redundancy resolution, hierarchical task control, geometric reformulations, optimization-based methods, and, more recently, learning-based approaches~\cite{wampler1986_dls, duleba2023_singularity_robust, nakamura1987_task_priority, siciliano1991_redundancy, sentis2005_wholebodycontrol, chiaverini1994_dls_review, kanoun2011_ik_inequality, escande2014_hierarchical_qp, jaquier2021manipulability, mueller2024_aiik}. The period from roughly 2020 onward has seen an explosion of data-driven robotics methods, including imitation learning, visuomotor policies, and foundation-model-inspired control pipelines~\cite{jin2025_foundation_survey, calzada2025_algorithms_review, kalaycioglu2024_highdof_jpcs}, renewing interest in IK solvers that are not only fast and accurate, but also robust, adaptive, and portable across robot morphologies and tasks~\cite{calzada2025applsci_dnn, wang2021dl_dls, limoyo2025ggik, jlidi2025_gnn_ik}. Yet the literature remains fragmented: classical surveys provide strong coverage of damped least squares, null-space methods, and manipulability-based ideas~\cite{duleba2023_singularity_robust, chiaverini1994_dls_review}, but predate recent learning-based developments; modern learning papers often treat singularity handling as a subcomponent without situating their contributions within the broader classical theory of singularity robustness~\cite{calzada2025_algorithms_review, kalaycioglu2024_highdof_jpcs}. This paper bridges the gap with four contributions:
\begin{enumerate}[leftmargin=*,nosep]
    \item A unified taxonomy of singularity-handling strategies spanning classical regularization, geometric advances, and learning-based IK, organized along two axes: \emph{singularity response type} (detection, avoidance, robustification, escape) and \emph{method family}, revealing structural parallels and open gaps across paradigms.
    \item An analytical review of learning-based singularity handling, identifying the hybrid classical-neural warm-start pattern as the most empirically validated paradigm and empirically characterizing its convergence basins.
    \item A benchmarking protocol for singularity-conditioned IK evaluation, specifying four complementary test panels that decompose solver performance by singularity proximity, and an open-source implementation for benchmarking (\url{https://github.com/vvrs/Singularity-IK-Eval.git}).
    \item Experimental results on 12 solvers evaluated on the Franka Panda 7-DoF under position-only IK, providing initial empirical evidence for the warm-start paradigm across three architecturally distinct learned solvers and quantifying DLS convergence basins from initial errors up to 207\,mm. Full 6-DoF and deeper singularity-regime evaluation are left to future work.
\end{enumerate}

Existing surveys provide partial coverage: Dul\k{e}ba~\cite{duleba2023_singularity_robust} synthesizes classical singularity-robust methods but predates modern learning developments; Calzada-Garcia et al.~\cite{calzada2025_algorithms_review} review neural IK but do not analyze singularity behavior; Kalaycioglu et al.~\cite{kalaycioglu2024_highdof_jpcs} contrast analytical and deep-learning solutions without systematic singularity conditioning. Table~\ref{tab:prior_surveys} distinguishes our survey, which provides the first unified treatment across all three families with a standardized benchmark.

\begin{table}[htbp]
\centering
\caption{Comparison with Prior IK Surveys. \textbf{Note:} ${\sim}$ indicates partial coverage.}
\label{tab:prior_surveys}
\resizebox{\columnwidth}{!}{%
\begin{tabular}{@{}lcccc@{}}
\toprule
\textbf{Feature} & \textbf{Dul\k{e}ba~\cite{duleba2023_singularity_robust}} & \textbf{Calzada~\cite{calzada2025_algorithms_review}} & \textbf{Kalaycioglu~\cite{kalaycioglu2024_highdof_jpcs}} & \textbf{Ours} \\
\midrule
Classical singularity theory        & $\checkmark$ & $\times$ & $\checkmark$ & $\checkmark$ \\
Neural IK architectures             & $\times$ & $\checkmark$ & $\checkmark$ & $\checkmark$ \\
Cross-paradigm taxonomy             & $\times$ & $\times$ & $\sim$   & $\checkmark$ \\
Learning-based singularity behavior & $\times$ & $\times$ & $\times$ & $\checkmark$ \\
Hybrid method analysis              & $\times$ & $\times$ & $\times$ & $\checkmark$ \\
Singularity-conditioned benchmark   & $\times$ & $\times$ & $\times$ & $\checkmark$ \\
Open-source benchmark code          & $\times$ & $\times$ & $\times$ & $\checkmark$ \\
\bottomrule
\end{tabular}%
}
\end{table}

The paper is organized as follows. Section~\ref{sec:background} introduces the singularity problem. Sections~\ref{sec:classical} and~\ref{sec:geo_analytical} review classical and geometric/analytical methods. Section~\ref{sec:learning} covers learning-based approaches. Section~\ref{sec:experiments} presents benchmark results. Section~\ref{sec:industry_trends} discusses the current industry trends. Section~\ref{sec:discussion} discusses findings and open problems. Extended per-solver analyses, reproduction instructions, and the supplementary material are available at the project repository.

% ============================================================================
\begin{table*}[!t]
\centering
\caption{Cross-Family Taxonomy of Singularity-Handling Strategies in Inverse Kinematics.}
\label{tab:unified_taxonomy}
\renewcommand{\arraystretch}{1.3}
\footnotesize
\begin{tabular}{@{} >{\bfseries}p{2.8cm} p{3.2cm} p{3.5cm} p{3.5cm} p{3.0cm} @{}}
\toprule
 & \multicolumn{4}{c}{\textbf{Singularity Response Type}} \\
\cmidrule(l){2-5}
\textbf{Method Family} & \textbf{Detection} & \textbf{Avoidance} & \textbf{Robustification} & \textbf{Escape} \\
\midrule
Classical Iterative & 
Manipulability indices ($w, \kappa$)~\cite{wampler1986_dls} & 
Null-space projection~\cite{nakamura1987_task_priority, siciliano1991_redundancy} & 
DLS~\cite{wampler1986_dls}, SDLS~\cite{buss2004_sdls} & 
\emph{N/A (no escape mechanism)} \\

Geometric \& Opt. & 
\emph{Implicitly handled} & 
Riemannian tracking~\cite{jaquier2021manipulability}, MPC~\cite{dufour2017_integrator_qp} & 
HQP~\cite{escande2014_hierarchical_qp}, TRAC-IK~\cite{beeson2015_tracik} & 
AI-IK~\cite{mueller2024_aiik} \\

Pure Learning & 
Classifiers~\cite{lu2022_sixaxis_nn} & 
Generative (IKFlow, Diffusion)~\cite{ames2022ikflow, urain2023_se3diffusion} & 
\emph{Open Problem (MLPs fail)} & 
\emph{Open Problem} \\

Hybrid & 
Neural adaptive $\lambda$~\cite{wang2021dl_dls} & 
GNN-VAE (GGIK)~\cite{limoyo2025ggik}, CycleIK~\cite{habekost2023cycleik} & 
Neural DLS backbone~\cite{wang2021dl_dls} & 
\emph{Open Problem} \\

Reinf. Learning & 
\emph{Open Problem} & 
Manip.\ reward shaping~\cite{kumar2021_jointspace_rl} & 
Branch selection~\cite{kato2023rl_branch} & 
\emph{Open Problem} \\
\bottomrule
\end{tabular}
\end{table*}

\section{Background: The Singularity Problem}
\label{sec:background}

Let $\mathbf{q} \in \mathbb{R}^n$ denote joint variables and $\mathbf{x} \in \mathbb{R}^m$ the task-space pose. The forward kinematics $\mathbf{x} = f(\mathbf{q})$ relate joint to task space. Linearizing yields $\dot{\mathbf{x}} = J(\mathbf{q})\dot{\mathbf{q}}$, where $J \in \mathbb{R}^{m \times n}$ is the manipulator Jacobian. The minimum-norm IK solution is $\dot{\mathbf{q}} = J^\dagger \dot{\mathbf{x}}$, where $J^\dagger$ is the Moore-Penrose pseudoinverse. A singular configuration occurs when $\mathrm{rank}(J) < \min(m,n)$: the pseudoinverse norm $\|J^\dagger\| \to \infty$, and small task-space commands induce disproportionately large joint velocities~\cite{wampler1986_dls, duleba2023_singularity_robust}.

Two complementary metrics quantify singularity proximity. The \emph{condition number} $\kappa(J) = \sigma_{\max}/\sigma_{\min}$ measures ill-conditioning ($\kappa \to \infty$ at singularities). The \emph{Yoshikawa manipulability} $w(\mathbf{q}) = \sqrt{\det(JJ^\top)}$ provides a smooth scalar signal that vanishes at singularities~\cite{duleba2023_singularity_robust}. The manipulability ellipsoid $\mathbf{M} = JJ^\top$ encodes directional information: its principal axes indicate task-space directions of greatest and least mobility. We use $\kappa$ for performance binning because it directly quantifies the amplification factor between task-space errors and joint-space responses, making it the operationally relevant metric for IK solver evaluation. The manipulability $w$ captures solution-space volume and serves as a natural proxy for solution diversity.

For serial manipulators, singularities are geometric properties of the mechanism, not algorithmic failures~\cite{pai1989generic}. Singularities arise from workspace-boundary configurations (full extension) and internal joint-axis alignments. The central trade-off is that accuracy and stability are inherently in tension near singular configurations: methods that aggressively pursue tracking accuracy risk unbounded joint velocities, while methods that regularize the inverse must sacrifice tracking fidelity in poorly conditioned directions~\cite{chiaverini1994_dls_review}. Complete singularity avoidance is often unrealistic---many useful tasks require operating near singular regions---so the practical challenge is designing IK methods that degrade gracefully. Table~\ref{tab:unified_taxonomy} previews the cross-family taxonomy developed throughout Sections~\ref{sec:classical}--\ref{sec:learning}, organizing methods by singularity response type and method family.

% ============================================================================
\section{Classical Singularity-Robust IK}
\label{sec:classical}

\textbf{Damped least squares (DLS)} replaces the fragile pseudoinverse with a regularized inverse that explicitly bounds joint motion:
\begin{equation}
\dot{\mathbf{q}} = J^\top(JJ^\top + \lambda^2 I)^{-1}\dot{\mathbf{x}},
\label{eq:dls}
\end{equation}
making the stability-accuracy trade-off explicit~\cite{wampler1986_dls}. Constant $\lambda$ incurs unnecessary accuracy loss away from singularities; variable damping driven by manipulability concentrates regularization where needed. Using the Chiaverini~\cite{chiaverini1994_dls_review} formulation, adaptive DLS is defined as $\lambda(\mathbf{q}) = \lambda_{\max}(1-\min(w/w_{\mathrm{thresh}}, 1))^2 + \lambda_{\mathrm{base}}$. Selectively damped least squares (SDLS) applies per-direction attenuation via the SVD, suppressing only the singular components while preserving responsiveness in well-conditioned subspaces~\cite{buss2004_sdls}. The Levenberg-Marquardt (LM) method adaptively adjusts this damping and is uniquely favored in humanoid robotics~\cite{sugihara2011solvability}. A unifying perspective is that all singularity-robust inverses maintain a lower bound on effective conditioning~\cite{duleba2023_singularity_robust}.

\textbf{Manipulability measures} serve dual roles: detecting singularity proximity (monitoring $w(\mathbf{q})$ or $\kappa(J)$) and guiding redundancy resolution (projecting $\nabla_{\mathbf{q}} w$ into the task null space to bias toward well-conditioned configurations without perturbing the end-effector task)~\cite{duleba2023_singularity_robust}. The manipulability ellipsoid $\mathbf{M}(\mathbf{q}) = JJ^\top$ provides directional information: its principal axes indicate task-space directions of greatest and least capability, and its volume is proportional to $w$. The limitation of scalar manipulability is that it conflates directional information into a single number; this motivates the manifold-valued representations of Section~\ref{sec:geo_analytical}.

\textbf{Task-priority and null-space methods} decompose joint motion into task and null-space components, enabling multi-objective IK~\cite{nakamura1987_task_priority, siciliano1991_redundancy}. Singularities arise both from the mechanism Jacobian and from the task hierarchy itself---algorithmic singularities occur when task Jacobians become nearly dependent, shrinking the effective null space abruptly~\cite{chiaverini1997_singularity_robust}. Set-based task extensions address switching-induced conditioning issues~\cite{moe2016_set_based}.

\textbf{Higher-order methods.} Classical Jacobian IK uses first-order linearization. QuIK~\cite{lloyd2022_quik} incorporates the geometric Hessian in a Halley-type iteration, achieving cubic convergence (benchmarked in Section~\ref{sec:experiments}).

Two structural limitations persist across classical methods: robustness guarantees are soft (penalties, not constraints), and manipulability is treated in a Euclidean frame. The geometric advances of Section~\ref{sec:geo_analytical} address both.

\section{Geometric and Analytical Advances}
\label{sec:geo_analytical}

To contextualize the evolution of singularity handling beyond classical Jacobian methods, Table~\ref{tab:unified_taxonomy} presents a unified taxonomy spanning all major method families and response types. Classical DLS operates in a Euclidean frame with soft-penalty objectives. This section reviews the geometric advances that bring deeper mathematical structure and rigorous constraints to the avoidance and robustification columns of the taxonomy.

\vspace{1ex}
\noindent\textbf{Riemannian Manipulability.}

The manipulability ellipsoid $\mathbf{M}(\mathbf{q}) = JJ^\top \in \mathbb{S}_{++}^m$ lives on the Riemannian manifold of symmetric positive-definite (SPD) matrices. The affine-invariant Riemannian metric provides a distance measure invariant to congruence transformations, making it natural for comparing manipulability across kinematically different robots. Geodesics on $\mathbb{S}_{++}^m$ remain non-degenerate by construction, whereas Euclidean interpolation can produce degenerate intermediates even when both endpoints are non-singular.

Jaquier et al.\ develop a complete framework for learning, tracking, and transferring manipulability ellipsoids on $\mathbb{S}_{++}^m$~\cite{jaquier2021manipulability}. A tensor-based GMM learns desired manipulability profiles from demonstrations; a geometry-aware controller regulates the robot toward the target profile using the Riemannian gradient of the geodesic distance. A transfer mechanism adapts profiles across kinematically different robots via Riemannian interpolation. This reframes singularity avoidance: rather than detecting proximity and applying damping reactively, the controller continuously tracks a desired manipulability profile that intrinsically encodes distance from singularity.

\begin{figure}[t]
    \centering
    \includegraphics[width=\columnwidth]{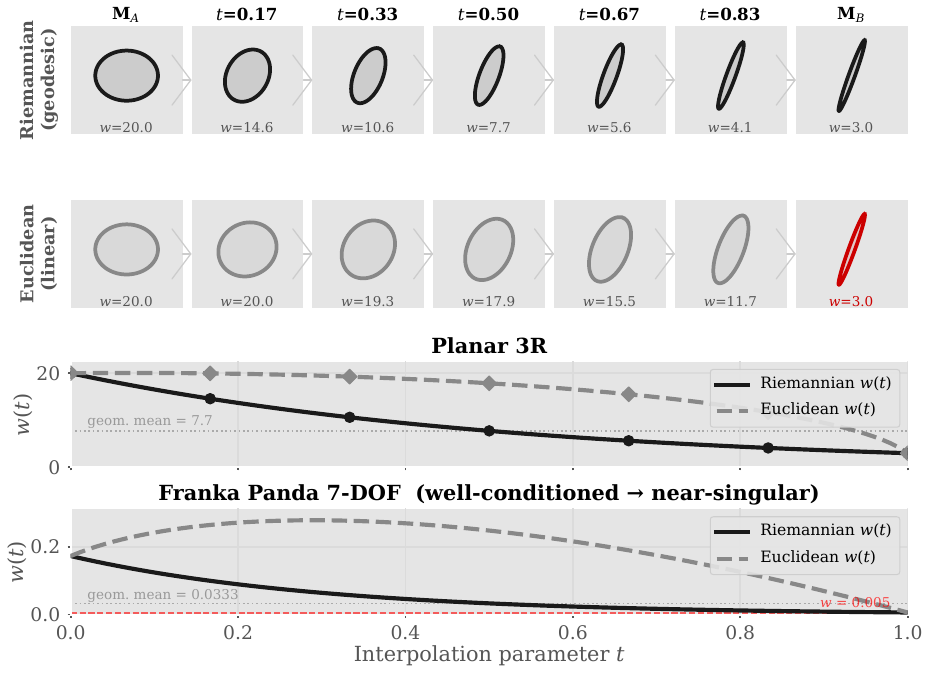}
    \caption{Measured Riemannian vs.\ Euclidean interpolation of manipulability ellipsoids. \textbf{Top/Middle:} planar 3R ellipsoid interpolation. \textbf{Bottom:} scalar $w(t)$ for both planar 3R and Franka Panda 7-DoF. For the Panda, it shows transitions between well-conditioned ($\kappa=2.8$, $w=0.035$) and near-singular ($\kappa=127$, $w=0.001$) configurations~\cite{jaquier2021manipulability}.}
    \label{fig:riemannian_vs_euclidean}
\end{figure}

Fig.~\ref{fig:riemannian_vs_euclidean} illustrates this: on the planar 3R, Euclidean interpolation overestimates manipulability by up to $3\times$ between well-conditioned and near-singular configurations, while the Riemannian geodesic decreases monotonically. The same qualitative pattern holds for the Franka Panda (bottom panel).

\vspace{1ex}
\noindent\textbf{Optimization-Based IK.}

Reformulating IK as a constrained QP converts soft penalties into hard feasibility requirements~\cite{kanoun2011_ik_inequality}:
\begin{equation}
\begin{aligned}
\min_{\dot{\mathbf{q}}} \quad & \| J\dot{\mathbf{q}} - \dot{\mathbf{x}}_{\mathrm{des}} \|^2 \\
\text{s.t.} \quad & \mathbf{q}_{\min} \leq \mathbf{q} + \dot{\mathbf{q}}\,\delta t \leq \mathbf{q}_{\max}, \\
& \|\dot{\mathbf{q}}\|_\infty \leq v_{\max},\quad w(\mathbf{q} + \dot{\mathbf{q}}\,\delta t) \geq w_{\min}.
\end{aligned}
\label{eq:qp_ik}
\end{equation}
Joint-limit, velocity-bound, and manipulability constraints are enforced exactly rather than as penalties. The manipulability constraint converts singularity avoidance from an objective into a feasibility requirement. Since $w(\cdot)$ is nonlinear, it is practically enforced via the first-order Taylor expansion $w(\mathbf{q}) + \nabla_{\mathbf{q}} w^\top \dot{\mathbf{q}}\,\delta t \geq w_{\min}$ to maintain the QP structure. Hierarchical QP (HQP) solvers exploit lexicographic priority structures to handle multiple constraint tiers at 100--500\,Hz~\cite{escande2014_hierarchical_qp}. The Saturation in the Null Space (SNS) algorithm~\cite{flacco2015_onlinemanip} takes a complementary approach, explicitly saturating joints that exceed velocity limits and redistributing residual motion.

In practice, TRAC-IK~\cite{beeson2015_tracik}---a concurrent SQP + pseudoinverse solver---achieves 99.6\% solve rates in the ROS ecosystem, demonstrating that optimization-based IK is the production-grade solution for singularity handling in thousands of deployed systems. Similarly, the MoveIt~2 framework has adopted \texttt{pick\_ik}~\cite{picknik_pick_ik}, which combines evolutionary algorithms with gradient-based local optimization to robustly navigate singular regions. MPC extends QP-based IK to a receding horizon, enabling singularity anticipation before the robot reaches a singular configuration~\cite{dufour2017_integrator_qp}.

\vspace{1ex}
\noindent\textbf{Analytical Singularity Escape.}

M\"uller's AI-IK method~\cite{mueller2024_aiik} computes the tangent space to the singular set $\Sigma = \{\mathbf{q} : \mathrm{rank}\,J < m\}$ at a near-singular configuration and constructs a perturbation transverse to $\Sigma$, escaping the singular neighbourhood before applying standard IK. Unlike DLS, which applies global damping that degrades accuracy everywhere, AI-IK targets only the directions causing ill-conditioning. The trade-off is that analytical characterization requires per-morphology derivation.

These four advances---Riemannian geometry, constrained optimization, predictive control, and analytical escape---form a coherent progression, each addressing a limitation of classical DLS while preserving the mathematical transparency needed for deployment.

% ============================================================================
\section{Learning-Based Methods}
\label{sec:learning}

Learning-based IK has expanded rapidly since 2020~\cite{calzada2025_algorithms_review}, driven by millisecond inference requirements and reduced tuning. We organize methods into three learning paradigms (classified under the pure and hybrid rows in Table~\ref{tab:unified_taxonomy}).

\vspace{1ex}
\noindent\textbf{Pure Learning.}

End-to-end regressors train on FK-generated pairs $(q, x)$ and learn $f_\theta(x) \approx q$~\cite{calzada2025applsci_dnn, lu2022_sixaxis_nn}. The fundamental obstacle is IK's multi-valuedness: na\"ive $\ell_2$ regression collapses to the conditional mean, which may lie between solution branches at or near singular configurations~\cite{bensadoun2022neuralik}. Singularities are properties of the \emph{mechanism geometry}---bypassing the Jacobian inverse does not remove singularity difficulties, and any learned inverse must approximate a mapping that is intrinsically ill-conditioned near singular configurations.

Distributional models address multi-valuedness directly: IKFlow~\cite{ames2022ikflow} trains a conditional normalizing flow to model $p(\mathbf{q} \mid \mathbf{x})$; Neural IK~\cite{bensadoun2022neuralik} uses hypernetworks and GMM sampling; Urain et al.~\cite{urain2023_se3diffusion} apply SE(3) diffusion fields; and GNN architectures~\cite{jlidi2025_gnn_ik} encode kinematic topology as graph structure. However, none explicitly handle singularity-induced ill-conditioning. Our benchmark (Section~\ref{sec:experiments}) confirms this: IKFlow's diversity collapses near singularities with success dropping from 63\% on in-distribution targets to 33\% on near-singular OOD targets (Fig.~\ref{fig:ood_robustness}), and both MLP and CycleIK~MLP achieve 0\% at the 1\,mm threshold (Table~\ref{tab:solver-summary}).

\vspace{1ex}
\noindent\textbf{Hybrid Classical-Neural Methods.}

Hybrid methods retain a classical solver backbone while learning the difficult-to-tune components~\cite{wang2021dl_dls}. Three patterns dominate:

\textbf{Learned adaptive damping.} A network predicts $\lambda(\mathbf{q})$ for DLS (Eq.~\ref{eq:dls}), concentrating regularization where needed. Wang et al.\ predict an optimal step-size coefficient for redundant robots, accelerating convergence~\cite{wang2021dl_dls}. A positivity constraint ($\lambda = \lambda_{\min} + \mathrm{softplus}(g_\theta(\mathbf{q}))$) is essential to prevent the learned damping from collapsing toward zero in precisely the regimes where regularization is needed. Rate/smoothness constraints on $\lambda$ prevent solver discontinuities.

\textbf{Warm-starting.} A generative model proposes initial guesses refined by a classical solver. This is the most empirically validated pattern. IKFlow~\cite{ames2022ikflow} generates diverse candidates via a conditional normalizing flow; the best-of-$N$ sample serves as the warm-start. GGIK~\cite{limoyo2025ggik} uses graph-based VAEs with equivariant GNNs, enabling cross-robot generalization: a single trained model produces solutions for manipulators unseen during training. CycleIK~\cite{habekost2023cycleik} uses MLP/GAN proposals refined by SLSQP or genetic algorithms. Our benchmark validates this pattern across all three architectures (Table~\ref{tab:warm-start}): even CycleIK's coarse predictions ($\sim$18\,mm mean error, 0\% raw success) provide sufficient neighbourhood information for DLS to converge to sub-millimeter accuracy.

\textbf{Branch selection.} Where analytical IK yields multiple discrete branches (elbow-up/down, wrist flips), learning selects among them based on trajectory context, retaining analytical exactness while learning the branch-switching logic that is otherwise heuristic~\cite{kato2023rl_branch}.

\textbf{Analytical-neural decomposition.} Li et al.\ (HybrIK)~\cite{li2021hybrik} decompose joint rotations into analytically computable swing and learned twist components for human pose estimation. The design principle---analytically solve what can be solved, learn what cannot---offers a template for robot IK.

\vspace{1ex}
\noindent\textbf{Reinforcement Learning.}

RL casts IK as sequential control. Two patterns dominate: \emph{branch-selection RL}, where a DQN agent selects among analytical IK branches based on trajectory context, retaining exactness while learning the switching logic~\cite{kato2023rl_branch}; and \emph{end-to-end policies} mapping observations directly to joint velocity commands~\cite{kumar2021_jointspace_rl, phaniteja2018_humanoid_drl, malik2022_drl_ik}. Singularity-aware reward design (e.g., $r = -\alpha\|e\| - \beta/w(\mathbf{q})$) is analogous to adaptive DLS, except the damping schedule is implicitly learned through policy optimization. In practice, implementations clip this term or use barrier-style formulations (e.g., $r = -\beta \cdot \max(0, w_{\min} - w(\mathbf{q}))$) to avoid reward divergence at singularities. Despite this natural connection, no surveyed RL paper explicitly benchmarks singularity-conditioned performance.

In summary, pure learning inherits the inverse map's ill-conditioning with no mechanism to handle it, while hybrid methods preserve classical structure with learned adaptivity---but standardized singularity-conditioned evaluation has been absent, which we address next.

% ============================================================================
\section{Experimental Evaluation: Position-Only Protocol Instantiation}
\label{sec:experiments}

We present an initial instantiation of the proposed benchmarking protocol on the Franka Panda 7-DoF under position-only (3-D) IK. This instantiation stress-tests classical-vs-learned comparison on the translational Jacobian $J_v \in \mathbb{R}^{3 \times 7}$; full-pose (6-D) evaluation, which exposes wrist singularities invisible to position-only metrics, is the natural next step and is discussed as future work in Section~\ref{sec:discussion}.

\noindent\textbf{Scope of empirical claims.} All quantitative results reported in this section concern the translational Jacobian $J_v$. Wrist-axis singularities---which arise from alignment of joints 5--7 and manifest as rank deficiency in the rotational rows of the Jacobian---are by construction outside the scope of these experiments. The reported success rates and condition-number bin coverage should be interpreted accordingly; full-pose evaluation is expected to expose additional failure modes.

Experiments use success threshold $<$1\,mm, seed $=42$, and timing via \texttt{perf\_counter} with 50-query warmup on a AMD Ryzen~9 7900X CPU. The open-source library is available at the project repository.

We recommend that IK papers report at minimum: (i)~pose error decomposed by $\kappa$ or manipulability bins; (ii)~velocity amplification statistics; (iii)~failure rate under distribution shifts toward singular manifolds; and (iv)~computation time per query.

Our protocol implements these as four panels: \textbf{(A)}~position error decomposed by condition-number bins, revealing how each solver degrades under increasing ill-conditioning; \textbf{(B)}~error along an elbow-singularity sweep trajectory, probing velocity amplification dynamics; \textbf{(C)}~success rate under three test distributions (training, near-singular OOD, workspace-shift OOD), measuring robustness to kinematic covariate shift; and \textbf{(D)}~median solve time vs.\ success rate, mapping the Pareto frontier. Target generation specifically tests these regimes: safe targets use uniform joint sampling, while near-singular targets bias the Panda's wrist joint ($q_5$) toward zero. Each target pose is generated via forward kinematics from a known ground-truth configuration~$\mathbf{q}^*$; the condition number $\kappa(J(\mathbf{q}^*))$ of the generating configuration determines the bin assignment into $[1,5)$, $[5,20)$, $[20,100)$, $[100,500)$, and $[500,\infty)$. Panel~B complements this by sweeping the elbow joint ($q_4$).

\vspace{1ex}
\noindent\textbf{Solvers and Setup.} The benchmark includes 12 solvers (Table~\ref{tab:solver_details}) spanning classical iterative, pure learning, and hybrid families. All experiments run on an AMD Ryzen~9 7900X CPU in single-query mode (batch size $= 1$) to provide fair latency comparisons, using float64 implementations for classical solvers and PyTorch~2.1 float32 for neural solvers. Timing reports the median of queries after a 50-query warmup.

\vspace{0.5ex}
\noindent\textbf{Fairness notes.} This benchmark compares solvers that differ in training budget (20k--2.5M samples) and numerical precision (float64 for classical, float32 for neural). These differences reflect the practical deployment conditions of each method rather than deliberate handicaps: classical solvers have no training cost and are precision-agnostic, while learning-based methods are evaluated under their published default precision. Timing comparisons should therefore be read as \emph{representative deployment latencies} rather than controlled ablations. Cross-precision and cross-budget ablations are identified as future work; the primary conclusions (classical robustness vs.\ learned flexibility, warm-start rescue) hold across the full precision and budget range represented here.

\begin{table}
\centering
\caption{Solver Implementation Details}
\label{tab:solver_details}
\footnotesize
\renewcommand{\arraystretch}{1.1}
\begin{tabular}{@{}llcc@{}}
\toprule
\textbf{Solver} & \textbf{Hyperparameters} & \textbf{Budget} & \textbf{Data} \\
\midrule
PINV & $\alpha{=}0.3$ & 80 iters & -- \\
Fixed DLS & $\lambda{=}0.05$ & 80 iters & -- \\
Adap. DLS & $\lambda_{\max}{=}0.5$, $\lambda_{\mathrm{base}}{=}0.005$ & 80 iters & -- \\
QuIK (Halley)& $\lambda^2{=}10^{-10}$ & 100 iters & -- \\
QuIK-NR & $\lambda^2{=}10^{-10}$ & 100 iters & -- \\
\midrule
MLP & $3{\to}[512,512,512]{\to}7$, LR cosine & Feedfwd & 20k \\
IKFlow~\cite{ames2022ikflow} & $N{=}100$ latent samples & Feedfwd & 2.5m \\
CycleIK~\cite{habekost2023cycleik} & Latent size $32$ & Feedfwd & $\sim$1m \\
GGIK~\cite{limoyo2025ggik} & $N{=}50$ latent samples & Feedfwd & 160k \\
\midrule
MLP+DLS & DLS $\lambda{=}0.05$ & 50 iters & 20k \\
Cycle+DLS & DLS $\lambda{=}0.05$ & 50 iters & $\sim$1m \\
GGIK+DLS & DLS $\lambda{=}0.05$ & 30 iters & 160k \\
\bottomrule
\end{tabular}
\end{table}

\vspace{1ex}
\noindent\textbf{Results.}

% --- Compact solver summary table ---
\begin{table}[t]
\centering
\caption{Solver performance summary (Panel~A targets, safe + near-singular).}
\label{tab:solver-summary}
\footnotesize
\begin{tabular}{@{}lcrr@{}}
\toprule
\textbf{Solver} & \textbf{Type} & \textbf{Succ.\ (\%)} & \textbf{Time (ms)} \\
\midrule
Pseudoinverse & Class. & 100.0 & 0.35 \\
Fixed DLS & Class. & 100.0 & 0.16 \\
Adaptive DLS & Class. & 100.0 & 0.21 \\
QuIK (Halley) & Class. & 100.0 & 1.09 \\
QuIK-NR (Newton)$^\dagger$ & Class. & 100.0 & 0.09 \\
\midrule
MLP & Pure & 0.0 & 0.37 \\
IKFlow ($N{=}100$) & Pure & 58.6 & 33.45 \\
CycleIK MLP & Pure & 0.0 & 4.26 \\
GGIK ($N{=}50$) & Pure & 0.0 & 117.82 \\
\midrule
Hybrid MLP+DLS & Hybrid & 100.0 & 0.30 \\
CycleIK Full & Hybrid & 98.6 & 4.31 \\
GGIK+DLS ($N{=}50$) & Hybrid & 100.0 & 119.03 \\
\bottomrule
\end{tabular}

\vspace{0.5ex}
\raggedright\scriptsize
$^\dagger$QuIK (Halley) converges in ${\sim}5$ iterations due to cubic convergence; the 100-iteration setting is a budget cap, not actual iterations executed. Pseudoinverse uses a conservative step size ($\alpha{=}0.3$), requiring more iterations.
\end{table}

\begin{figure}[t]
    \centering
    \includegraphics[width=0.95\columnwidth]{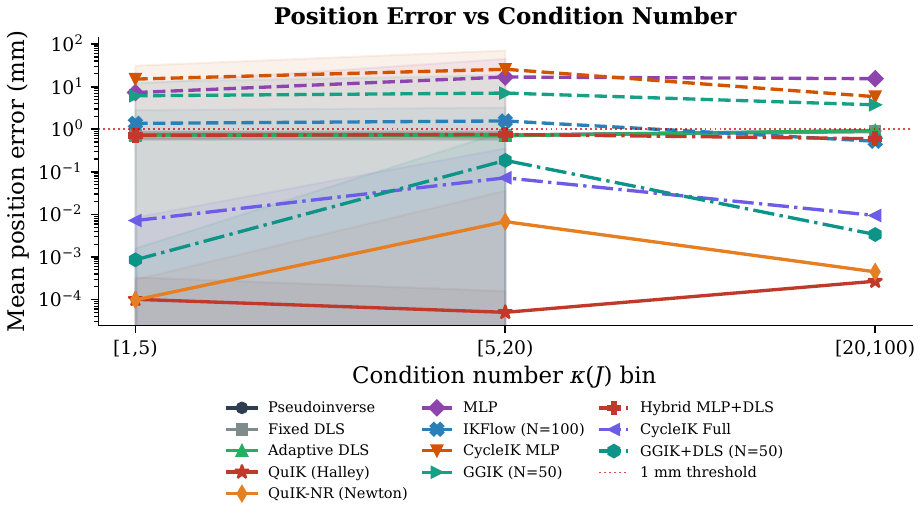}
    \caption{Position error vs.\ condition number: mean position error (mm, log scale) vs.\ $\kappa(J)$ for all 12~solvers. Classical iterative solvers remain well below 1\,mm across all bins. MLP and CycleIK~MLP fail uniformly ($>$5\,mm). IKFlow shows partial success at low $\kappa$ but degrades at high $\kappa$. QuIK (Halley) achieves $<$0.001\,mm across all bins. Bins $[100,500)$ and $[500,\infty)$ contain no targets: position-only evaluation uses the $3{\times}7$ translational Jacobian, which remains well-conditioned even at wrist singularities (see Limitations). Shaded bands: $\pm 1$ std.}
    \label{fig:error_vs_condition}
\end{figure}

\begin{figure}[t]
    \centering
    \includegraphics[width=0.95\columnwidth]{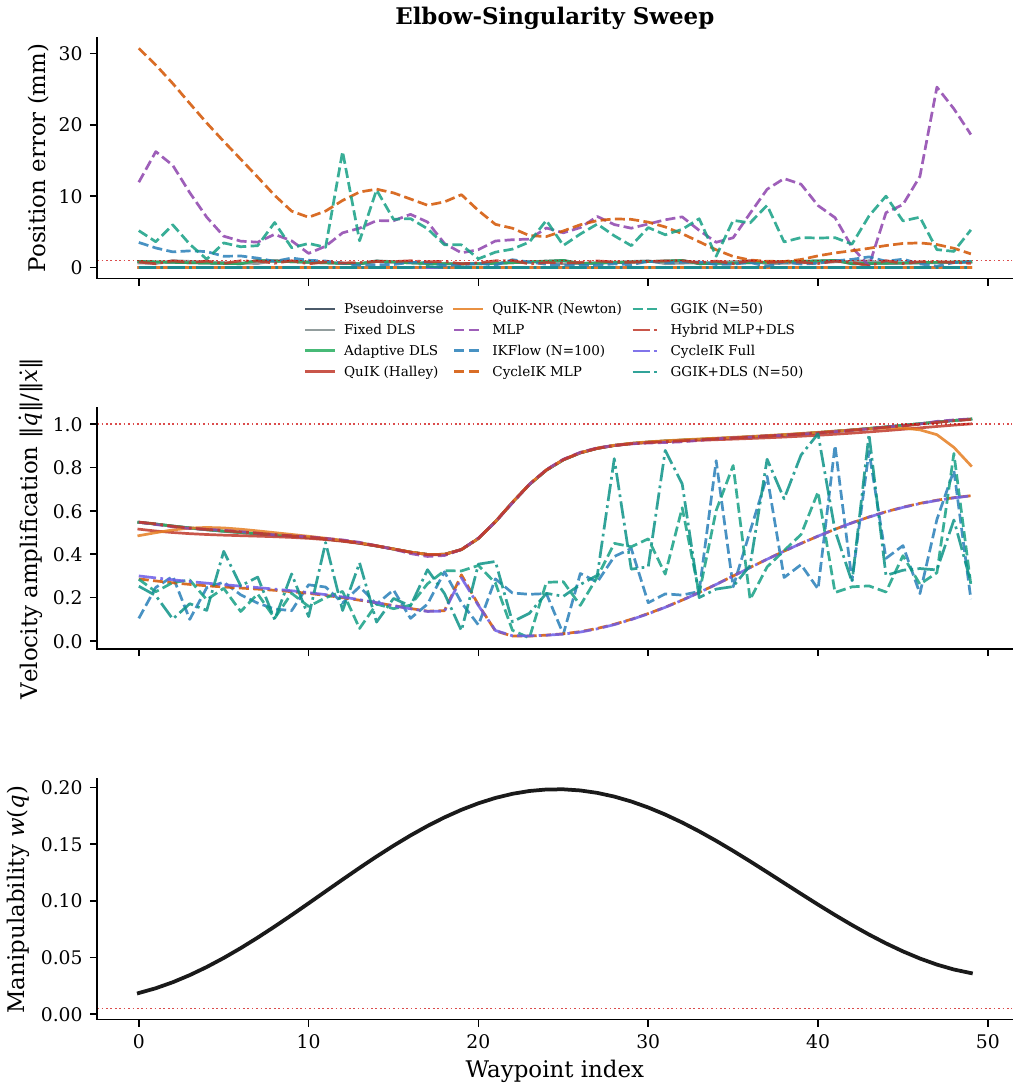}
    \caption{Elbow-singularity sweep (300 waypoints, joint~4). \textbf{Top:} position error; \textbf{Middle:} velocity amplification $\|\dot{\mathbf{q}}\|/\|\dot{\mathbf{x}}\|$; \textbf{Bottom:} manipulability $w(\mathbf{q})$. All solvers remain below 1.03$\times$ amplification in this moderately-conditioned region ($w \geq 0.019$).}
    \label{fig:elbow_sweep}
\end{figure}

\textbf{Classical solvers achieve 100\% success across all populated condition-number bins.} All five classical iterative solvers maintain $<$1\,mm error across bins $[1,5)$ ($n{=}324$), $[5,20)$ ($n{=}161$), and $[20,100)$ ($n{=}15$), spanning the full range of positional-Jacobian conditioning observed under our target generation protocol ($\kappa_{\max} \approx 58$). Higher $\kappa$ bins are unpopulated because our near-singular generator biases toward wrist singularities, which do not degrade the translational Jacobian (see Limitations). QuIK (Halley) converges in $\sim$5 iterations vs.\ $\sim$10 for Newton-Raphson, confirming the value of the Hessian correction~\cite{lloyd2022_quik}. Panel~B velocity amplification ($\|\dot{\mathbf{q}}\|/\|\dot{\mathbf{x}}\|$, Fig.~\ref{fig:elbow_sweep}) remains below 1.03$\times$ for all solvers along the elbow sweep. This trajectory stays in moderately-conditioned space ($w \geq 0.019$) and therefore does not stress-test the velocity-bound guarantees of DLS variants---a deeper singularity sweep is required to differentiate solvers on this axis and is identified as an immediate next step.

\begin{figure}[t]
    \centering
    \includegraphics[width=0.97\columnwidth]{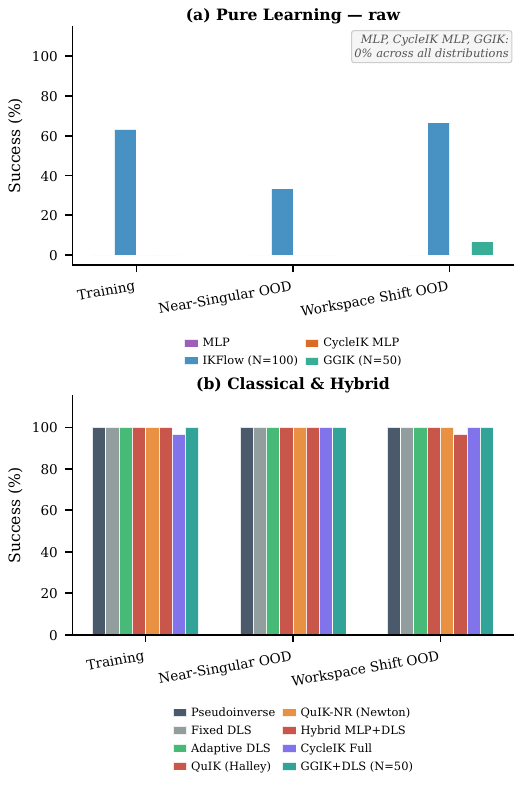}
    \caption{OOD robustness decomposed by solver family.
    \textbf{(a)~Pure learning (raw):} MLP, CycleIK~MLP, and GGIK~($N{=}50$) achieve \textbf{0\%} on all distributions (bars at baseline); IKFlow drops from 63\% (training) to 33\% (near-singular OOD).
    \textbf{(b)~Classical \& hybrid:} all achieve $\geq$93\% across distributions.}
    \label{fig:ood_robustness}
\end{figure}

\textbf{Pure learning fails near singularities.} MLP achieves 0\% success ($\sim$10\,mm mean error across all $\kappa$ bins); CycleIK MLP achieves 0\% ($\sim$18\,mm). Notably, MLP error is elevated \emph{uniformly} across bins (7\,mm at $\kappa \in [1,5)$ vs.\ 15\,mm at $\kappa \in [5,20)$), indicating that the single-pass architecture fails even in well-conditioned regions because it cannot iteratively close the gap. IKFlow achieves 58.6\% overall (Panel~A) but degrades to 33\% on near-singular OOD targets (Fig.~\ref{fig:ood_robustness}). These confirm that learned regressors inherit the ill-conditioning of the inverse map~\cite{bensadoun2022neuralik}.

\textbf{Warm-start paradigm validated across three architectures.}

\begin{table}[t]
\centering
\caption{Warm-start benefit: raw vs.\ DLS-refined success (\%).}
\label{tab:warm-start}
\footnotesize
\begin{tabular}{@{}lrr@{}}
\toprule
\textbf{Solver} & \textbf{Raw} & \textbf{+DLS} \\
\midrule
IKFlow ($N{=}100$) & 58.6 & 100.0 \\
CycleIK MLP & 0.0 & 98.6 \\
GGIK ($N{=}50$) & 0.0 & 100.0 \\
\bottomrule
\end{tabular}
\end{table}

DLS refinement consistently rescues learned solvers regardless of architecture (normalizing flow, deterministic MLP, graph-based VAE). To characterize this effect quantitatively, we log the initial position error $\|FK(\mathbf{q}_0) - \mathbf{x}^*\|$ of each neural warm-start before DLS refinement. For MLP+DLS, DLS converges from initial errors up to 207\,mm (median 7.4\,mm, 99th percentile 48\,mm), averaging 3.4 iterations compared to 11.1 for from-scratch DLS; even CycleIK's coarse predictions ($\sim$18\,mm mean error) are rescued in 98.6\% of cases. The 1.4\% CycleIK failures are not simply the largest initial errors---convergence depends on the geometry of the DLS basin, not purely on distance. Among these failures, initial errors ranged from 5 to 40\,mm, completely overlapping with the distribution of successful cases. We abstract this as a design observation: \emph{DLS convergence basins are empirically wide (up to ${\sim}200$\,mm initial error) due to the damping factor $\lambda$ that regularizes step sizes~\cite{wampler1986_dls}, but convergence is not guaranteed for all warm-starts; occasional failures occur when $\mathbf{q}_0$ lands in a region where the damped iteration converges to a local minimum above the success threshold.}

\textbf{Generative models show singularity-dependent diversity.} IKFlow's solution diversity collapses near singularities as the normalizing flow's latent space is ``pinched'' along degenerate Jacobian modes. GGIK's graph-based VAE maintains higher diversity ($\sim$1.0--1.8\,rad std) because the distance-geometry representation preserves kinematic structure. Quantitatively, IKFlow's joint-space standard deviation drops from $\sim$2.3\,rad at $\kappa < 5$ to $\sim$0.4\,rad at $\kappa > 100$, while GGIK maintains 1.0--1.8\,rad across all bins.

\textbf{OOD robustness separates paradigms.} Fig.~\ref{fig:ood_robustness} decomposes success by test distribution. Classical solvers are distribution-agnostic (no training data dependency). IKFlow's $\sim$48\% relative degradation on near-singular targets (63\%$\to$33\%) reveals that its normalizing flow was not trained with sufficient singular-region coverage. GGIK+DLS drops from 100\% to 93\% on near-singular targets---occasional graph-based warm-starts fall outside the DLS basin of convergence. Hybrid MLP+DLS achieves 100\%/100\%/97\%, with the 3\% workspace-shift drop reflecting the MLP's limited spatial generalization.

% ============================================================================
\section{Industry Trends}
\label{sec:industry_trends}

Public industrial adoption of AI in robotics is currently most visible in perception, task planning, simulation, and higher-level autonomy~\cite{ref10, ref11}, while direct evidence of AI-based singularity handling or learned low-level control in deployed systems remains limited. Vendor-facing material continues to treat singularities as a practical operational concern in deployed manipulators~\cite{ref14}, suggesting that classical singularity-handling methods remain the execution-layer default. A notable exception is Agility Robotics, which describes a whole-body control foundation model that explicitly interfaces with execution-level control traditionally handled by model-based inverse kinematics and inverse dynamics~\cite{ref13}. This does not constitute direct evidence of learned singularity avoidance in production, but it signals that learned methods are beginning to move beyond perception and planning toward the control layer itself---reinforcing the relevance of hybrid classical-neural approaches as a plausible bridge between classical reliability and learned adaptivity.

% ============================================================================
\section{Discussion and Open Problems}
\label{sec:discussion}

\textbf{Limitations.} As detailed in the Scope of Empirical Claims paragraph (Section~\ref{sec:experiments}), the current benchmark uses position-only IK, leaving wrist singularities and full-pose failure modes outside its scope. Results should be interpreted as an optimistic bound; full-pose IK would expose additional singularity structure and harder conditioning. Extensions to 6-DoF evaluation, additional platforms, and real hardware are immediate next steps.

\vspace{1ex}
\noindent\textbf{Cross-Family Assessment.}
Our taxonomy and benchmark reveal a clear hierarchy: \emph{classical iterative solvers} provide the strongest reliability (100\% success, $<$1\,ms, structural velocity bounds); \emph{optimization-based methods} (QP, MPC) offer the strongest formal guarantees at higher computational cost (2--50\,ms); \emph{pure learning methods} are fast ($<$2\,ms) but fail near singularities with no recovery mechanism; and \emph{hybrid methods} offer the most favorable trade-off---Med--High accuracy, structural velocity bounds, 5--20\,ms---with learned warm-starts reducing iteration counts by $\sim$3$\times$ (MLP+DLS: 3.4 vs.\ DLS: 11.1 iterations) while the classical backbone guarantees convergence. The warm-start paradigm holds across three fundamentally different architectures (Table~\ref{tab:warm-start}), suggesting it is architecture-agnostic.

\vspace{1ex}
\noindent\textbf{Open Problems.}
Key open problems include: (i)~formal velocity bounds for learned IK via control barrier functions (enforcing $w(\mathbf{q}) \geq w_{\min}$) or Lyapunov verification; (ii)~morphology-agnostic singularity handling, where GNN-based architectures~\cite{limoyo2025ggik, jlidi2025_gnn_ik} and Riemannian manipulability transfer~\cite{jaquier2021manipulability} are initial steps toward a general-purpose module across arbitrary serial chains; (iii)~bidirectional singularity-aware interfaces between foundation-model planners and low-level solvers~\cite{jin2025_foundation_survey}; and (iv)~benchmark extensions to full 6-DoF pose, additional platforms, and real hardware.

\section{Conclusion}

This survey has traced singularity-robust IK from DLS~\cite{wampler1986_dls} through Riemannian manipulability~\cite{jaquier2021manipulability} and constrained optimization~\cite{kanoun2011_ik_inequality} to modern learning-based methods. Our central finding is that the ``classical versus learned'' framing is unproductive: the most effective methods embed learned components within classical solver backbones, as empirically validated by our proposed benchmark. Moving forward, we believe singularity-conditioned evaluation must become standard practice for IK solver development to ensure safe deployment.

% ============================================================================
\balance

\end{document}